# 3D Facial Matching by Spiral Convolutional Metric Learning and a Biometric Fusion-Net of Demographic Properties

Soha Sadat Mahdi[1]*, Nele Nauwelaers[1]*, Philip Joris[1], Giorgos Bouritsas[2], Shunwang Gong[2], Sergiy Bokhnyak[3], Susan Walsh[4], Mark D. Shriver[5], Michael Bronstein[2,3,6], Peter Claes[1,7]

[1]KU Leuven, ESAT/PSI - UZ Leuven, MIRC; [2]Imperial College London, Department of Computing; [3]USI Lugano, Institute of Computational Science; [4]Indiana University-Purdue University-Indianapolis, Department of Biology; [5]Penn State University, Department of Anthropology; [6]Twitter; [7]KU Leuven, Department of Human Genetics

*these authors contributed equally and are ordered alphabetically

*Abstract*—Face recognition is a widely accepted biometric verification tool, as the face contains a lot of information about the identity of a person. In this study, a 2-step neural-based pipeline is presented for matching 3D facial shape to multiple DNA-related properties (sex, age, BMI and genomic background). The first step consists of a triplet loss-based metric learner that compresses facial shape into a lower dimensional embedding while preserving information about the property of interest. Most studies in the field of metric learning have only focused on 2D Euclidean data. In this work, geometric deep learning is employed to learn directly from 3D facial meshes. To this end, spiral convolutions are used along with a novel mesh-sampling scheme that retains uniformly sampled 3D points at different levels of resolution. The second step is a multi-biometric fusion by a fully connected neural network. The network takes an ensemble of embeddings and property labels as input and returns genuine and imposter scores. Since embeddings are accepted as an input, there is no need to train classifiers for the different properties and available data can be used more efficiently. Results obtained by a 10-fold cross-validation for biometric verification show that combining multiple properties leads to stronger biometric systems. Furthermore, the proposed neural-based pipeline outperforms a linear baseline, which consists of principal component analysis, followed by classification with linear support vector machines and a Naïve Bayes-based score-fuser.

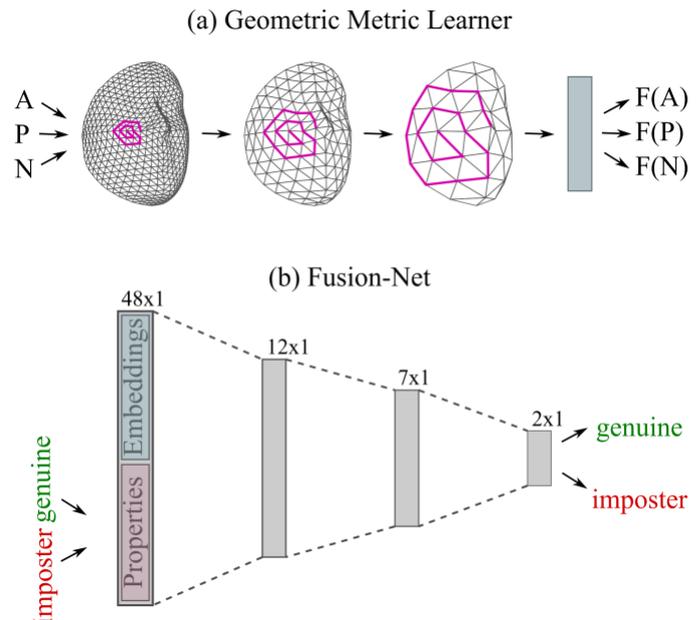

Figure 1: 2-Step neural pipeline for matching 3D facial images to DNA-related properties. (a) Geometric metric learner that takes anchor A, positive P and negative N facial meshes as input and generates corresponding embedding vectors. (b) Fusion-Net that takes embeddings and DNA-related properties and predicts genuine and imposter matching scores.

## I. INTRODUCTION

The human face contains substantial information about many genetic or environmental factors related to the identity of a person, which is useful in a biometric system. Facial recognition from DNA refers to a biometric system that aims to identify or verify DNA-related traits against facial images with known identity. With advances in the field, DNA phenotyping has attracted the attention of forensic biologists and anthropology researchers [1]. However, directly predicting faces from DNA remains an unsolved problem. Recently, Sero et al. [2] proposed a state-of-the-art approach which incorporates face-to-DNA classifiers, in order to classify 3D facial images by a group of demographic properties. These properties can either be decoded directly - sex, genomic background (GB) - or inferred indirectly – Body Mass Index (BMI), age - from DNA. Subsequently, the classification scores for all properties individually are fused by a Naive Bayes score fuser into a single matching score for each {face, list of properties} tuple. This approach involves three steps: 1) unsupervised dimensionality reduction of faces, 2) training of classifiers to obtain classification scores, 3) training a score fuser. It requires three independent data partitions: One for training the individual classifiers, one for training the score fuser and, lastly, one for testing the system.

In this paper, a 2-step neural-based pipeline for 3D data is proposed as an alternative, which is efficiently trainable with fewer data partitions, and, at the same time, improves the performance of the existing biometric system. The first stage of the proposed pipeline combines state-of-the-art geometric deep learning with metric learning techniques to obtain a semantic low dimensional embedding space for 3D facial data. In the second stage, a neural alternative to the traditional linear classification-based fuser is introduced. This neural-based fuser directly fuses information from the metric embeddings, which eliminates the need for an intermediate classification. An extended version of this introduction is available at [3].

## II. RELATED WORK

### A. Metric Learning

In many computer vision problems, projecting raw data into a compact yet meaningful space is crucial. Metric learning refers to the task of learning a semantic representation of data, based on the similarity measures defined by optimal distance metrics [4]. With the recent success of deep learning techniques, researchers of the field have introduced multiple deep metric learning tools [5]–[12]. Among those commonly used for face recognition, verification or person re-identification [14]–[19], are Siamese [19] and Triplet networks [20], which are twin or triple architectures with identical subnetworks. Some studies [21]–[24] suggest that the combination of a classification and verification loss can have a superior performance specifically for person re-identification. However, [25] claims that triplet loss and its variants outperform most other published methods by a large margin. The importance of a proper triplet mining strategy is a topic of discussion in the literature. [20] uses an online mining strategy of the negative samples that ensures a consistently increasing difficulty throughout the training within a face recognition system. In [7], however, a moderate mining for person re-identification is incorporated. Inspired by the literature, in this work, triplet architectures are adopted to learn compact embeddings of 3D facial meshes.

### B. Geometric Deep Learning

Geometric Deep Learning is a term that refers to methods that are designed for applying deep learning techniques onto non-Euclidean domains such as graphs and manifolds [26]. The application of deep learning methods for non-Euclidean data poses several problems, however. The first challenge is in defining a convolutional operator for graphs or meshes. Within geometric deep learning, there are two tracks to address this problem. On the one hand, spectral methods [27] involve those that are based on the frequency domain. The main drawbacks of these methods for shape analysis are that 1) the filters are basis dependent and can vary significantly for small perturbations on the shape, and 2) that there is no guaranteed spatial localization of the filters. However, these drawbacks can be tackled by spectrum-free approaches [27][28] that represent the filters via a polynomial expansion instead of operating explicitly in the frequency domain [26]. On the other hand, spatial methods [29][30] define a local system of coordinates along with a set of weighting functions. This results in a patch-operator that can be applied to each vertex of the graph or manifold. The second challenge is defining a pooling operator. Several graph coarsening or mesh decimation methods that are suitable for this task were proposed by [32]–[34].

In [35] an elegant and simple spatial convolution operator in the shape of a spiral defined on a mesh is introduced. This spiral acts as an anisotropic filter that slides over the mesh similar to convolutional filters on Euclidean domains. In [36] this technique was successfully applied to develop a generative model that is able to learn 3D deformable shapes with fixed topology such as facial and body scans. In this work, we extend the use of this approach to discriminative models that are trained to differentiate between facial shapes based on DNA-related soft traits or properties.

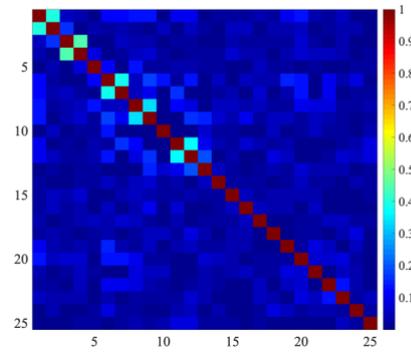

Figure 2: Correlation matrix for genomic background components

### C. Biometric Fusion

A biometric recognition system is described as a pattern recognition system that is designed to identify a person based on human characteristics [37]. These systems are generally based on a single biometric cue. Uni-biometric systems only consider a single biometric trait (e.g. face, fingerprint, iris) observed with a single method. To increase reliability, multi-biometrics, also referred to as biometric fusion, are introduced in which multiple inputs to the biometric system are accepted. The goal of the fusion is to obtain an overall system that is less prone to errors as opposed to each uni-biometric system. To achieve this, a fuser has to efficiently merge different inputs to eliminate their individual shortcomings and highlight their strengths [38]. When developing a multi-biometric system, two main design choices are presented: (i) what to fuse, and (ii) when to fuse [39]. Choice (i) refers to the selection of different input sources. There are multi-sensor [40], multi-algorithm [41], multi-instance [42], and multi-sample [43] systems, where different occurrences of the same trait are presented. In addition, multi-modal systems [44]–[46] exist where different biometric traits are fused in one system. Choice (ii) refers to the level in the biometric pipeline at which the fusion will take place. Examples are sensor-level [47], feature-level [48], score-level [49], rank-level [50] and decision-level [51] fusion.

### D. Face-to-DNA matching

In recent work [2], Sero et al. introduced a novel approach for matching between different identifiers (facial shape and DNA). Multiple face-to-DNA classifiers were trained, followed by a classification-based score-level fusion. The face was analyzed in a global-to-local way by performing a hierarchical segmentation of the facial shape. For each segment, a PCA model was built to construct unsupervised multi-dimensional shape features. Association studies based on canonical correlation analysis were performed to investigate the correlation between each of the segments and the following DNA-related properties: sex, age, BMI, and genomic background represented by 987 principal components. Significant segments were then used to train binary SVM classifiers. Continuous properties were binarized by applying a threshold: 30 years old for age, 23.62 kg/m$^2$ for BMI and zero for each of the genomic background components. The outputs of these classifiers are matching scores for all traits. Finally, the scores of different traits are fused into an overall matching score using a classification-based Naïve Bayes biometric fuser. This work has proven that matching different identifiers can lead to a successful biometric system, hereby avoiding the need to directly predict facial shape from DNA. Inspired by this work, we implemented a pipeline that has the same goal of matching facial shape to DNA by feature extraction and biometric fusion. However, in this work a fully neural-based pipeline is presented instead.

### III. MATERIALS AND METHODS

#### A. Dataset

The dataset used in this paper originated from studies at Pennsylvania State University (PSU) and Indiana University-Purdue University Indianapolis (IUPUI). A total sample size of n = 2,145 is used. The dataset includes texture-free 3D facial images, self-reported properties such as sex, age and BMI at the time of the collection, and genotypic data from individuals. The majority are female (68%), the age range is from 5 years old to 80 years old with an average of 27.39 years, and the BMI ranges from 11.87 kg/m$^{-2}$ to 62.11 kg/m$^{-2}$ with an average of 25.03 kg/m$^{-2}$. Recruited individuals are genetically heterogeneous, which implies that they originate from different background populations and admixtures thereof (e.g. European, Afro-American).

Study participants were sampled under Institutional Review Board (IRB) approved protocols (PSU IRB #44929, #45727, #2503, #4320, #32341 and IUPUI #1409306349) and were genotyped by 23andMe (23andMe, Mountain View, CA) on the v4 genome-wide SNP array and on the Illumina Multi-539 Ethnic Global Array (MEGA), respectively. Genotypes were imputed to the 1000 Genomes (1KGP) Project Phase 3 reference panel. SHAPEIT2 was used for prephasing of haplotypes and the Sanger Imputation Server PBWT pipeline was used for imputation. Using approximately 3.7M SNPs, a 25-dimensional SUGIBS space [52] was constructed based on the genetic data from the 1KGP. Subsequently, the participants from our dataset were projected onto the 1KGP components (Fig. 3). This resulted in an array of 25 components for each participant representing genomic background.

The 3D images were captured using 3DMD or Vectra H1 3D imaging systems. Participants were asked to close their mouths and keep a neutral expression. A spatially dense registration is performed on the images using MeshMonk [53]. Images are purified by removing hair and ears. Afterwards, five key landmarks are roughly positioned on the corners of the eyes, nose tip and mouth corners to guide a rigid surface registration of an anthropometric mask using the Iterative Closest Point (ICP) algorithm. Then, the mask, that consists of 7,160 quasi-landmarks, is non-rigidly registered to the faces, using non-rigid ICP, which leads to obtaining meshes with the same topology across the dataset. Meshes are then symmetrized by averaging them with their reflected image. Finally, Generalized Procrustes Analysis is performed to eliminate differences in scale, position and orientation [54].

#### B. Pipeline

The proposed pipeline, shown in Fig. 1, consists of two main steps: (I) Geometric Metric Learning (GML), for extracting a compact embedding from 3D facial meshes, that reflects the prominent information with respect to each property; and (II) Fusion-Net, for a biometric fusion of the embeddings and DNA-related properties into a single matching score. Such a fused score measures how well an individual's face matches to a list of given properties as extracted from DNA. A biometric verification then occurs if the score is deemed high enough.

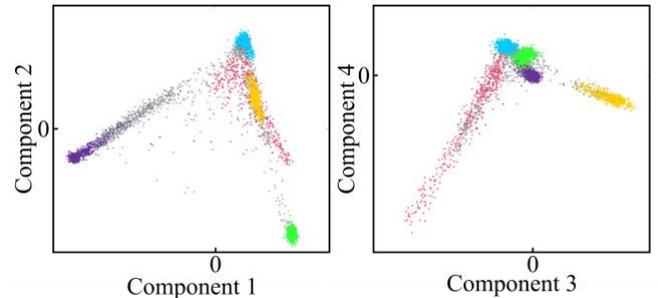

Figure 3: Scatterplot showing the first four GB components for our dataset (gray) and labeled 1KGP reference dataset. Populations: EUR = European, EAS = East Asian, AMR = Ad Mixed American, SAS = South Asian, AFR = African

##### 1) Geometric Metric Learning

For sex, age, and BMI, a separate triplet network is trained to encode 3D facial meshes into their corresponding low dimensional embedding vectors (Fig. 1.a). Later on, a concatenation of these embeddings and a property vector will form the input to the next step. Triplet networks are trained with triplets of (anchor, positive, negative), in which anchor and positive are samples from the same class, while the anchor and negative are from different classes. However, this definition is not easily applicable to properties represented by continuous values (age and BMI). Inspired by Jeong et al. [55], for any anchor, samples are considered positive when their label is within a certain distance from the label of the anchor, and negative otherwise. The distance thresholds T for Age and BMI are 10 and 2 respectively, which are selected after experimenting with multiple values between [1:20] for age and [1:5] for BMI. GB is the only multi-component property, and the correlation between different components is illustrated by the correlation matrix (Fig. 2). In order to take this correlation into account, a single triplet network is trained for all 25 components jointly. In each training batch, triplets are generated with respect to one specific component. This component is selected according to a weighted random number generator, where weights are the inverse of their prediction accuracy estimated from the previous epoch. To simplify the task, the GB components were transformed to a binary vector by a threshold of zero. Selecting a proper triplet mining strategy can have a significant impact on efficiency and accuracy of the training. After experimenting with different triplet mining strategies, random mining showed superior performances for all properties.

The design choice of the triplet subnetworks has a significant impact on obtaining optimal embeddings. Since our dataset consists of 3D facial meshes, regular deep learning techniques are not directly applicable. One option suggested in literature is to transform the data to the Euclidean domain and use a UV- or voxel-representation [56]–[58]. However, these transformations often involve a loss of data quality. Another option is to learn from the point cloud as in [59], but since this approach ignores the local connectivity of the mesh, it also comes with significant information loss. Instead, we use geometric deep learning to learn directly from the 3D facial meshes and optimally utilize the data structure. Spatial convolution operators show superior performance for tasks where spatial localization is desired. The spiral convolution is used as a convolution operator, since, in contrast to most traditional graph neural networks, it produces anisotropic filters, and it is especially powerful for shapes represented on a fixed topology. For each vertex of the mesh, a spiral is computed so as to contain vertices that are at most two hops away from the center vertex (Fig. 1.a). This spiral will operate similar to a standard convolution filter. Regular 1D convolutions can be expressed as

$$(f * g)_i = \sum_{j=0}^{J-1} f_{i-j} g_j + b \quad (1)$$

where f is the signal to which filter g with size J is applied and b is an added bias. For the spiral convolution, a weight is assigned to each point of the spiral, and using these weights the spiral can act as a filter applied to each point of the mesh:

$$(f * g)_i = \sum_{j \in spiral_i} f_j g_j + b \quad (2)$$

The weights of the spiral filters are learned by the network in the backpropagation step.

Aside from the convolution operator, a pooling operator for meshes must be introduced. Until now, most implementations for geometric deep learning rely on established mesh decimation techniques that reduce the number of vertices such that a good approximation of the original shape remains, but they result in irregularly sampled meshes at different steps of resolution. We, however, developed a 3D Mesh down- and up-sampling scheme that retains the property of equidistant mesh sampling. This is done so under the assumption that convolutional shape filters might benefit from the constant vertex density along the surface, such that they become more equally applicable on different regions of the shapes under investigation. This should improve the learning process, alongside the generalizability of the shape filters. The sampling scheme is based on the remeshing technique proposed in [60] and is computed by the following steps (Fig. 4)

i. The 3D mesh is mapped to a 2D unit square by means of a conformal mapping [61][62]. The boundary constraints forces all vertices at the boundary of the mesh to be mapped to one of the sides of the unit square.

ii. After the UV mapping, the area for highly detailed regions, such as the nose, is sampled more densely. To avoid losing information in these regions, the points are redistributed over the plane. A vector field is created to move the points towards a better distribution. Points lying close to each other, will push their neighbors away, while points far away from each other will pull their neighbors closer. For every point $P_i$, we compute the forces $F_{ij}$ it will apply to all other points $P_j$. The direction of $\vec{F}_{ij}$ is the direction of vector $\overrightarrow{P_i P_j}$. The magnitude of $\vec{F}_{ij}$ depends on two factors: the distance $d_{ij}$ between $P_i$ and $P_j$ and the distance of $P_i$ to its immediate neighbors. For the first factor, a weight $w_{ij} = e^{-d_{ij}/\sigma}$ is assigned, where $\sigma = 0.5$ for the first iteration and diminishes with 5% for following iterations. The second factor is based on the area $\Delta_i$, which is the sum of the area of the triangles surrounding $P_i$. A Gaussian weighting is applied so that points whose area $\Delta_i$ diverges a lot from the average area $\Delta_{avg}$ generate stronger forces. A pushing force is applied when $\Delta_i < \Delta_{avg}$, while a pulling force is applied by points for which $\Delta_i > \Delta_{avg}$. The vector field VF is the result of averaging all forces applied to each point: $VF_j = \sum_{i=1}^{n} \frac{F_{ij}}{n}$. The process of generating and applying vector fields is repeated for 12 iterations until the desired equidistant distribution is reached.

iii. The irregular 2D mesh is transformed to the Euclidean domain by interpolation with barycentric coordinates, resulting in three arrays containing x-, y- and z-values for each point.

iv. New meshes at different levels of resolution are constructed by defining a low-resolution base mesh which is then further refined. The base mesh consists of five vertices: four vertices at the corners of a unit square and one central vertex that is placed at the tip of the nose in the output of step (ii). Each side of the square serves as an edge and all corners are connected to the central vertex. The refinement is done with loop subdivision [63] by splitting each triangular face of the mesh into four smaller triangles by connecting the midpoints of the edges.

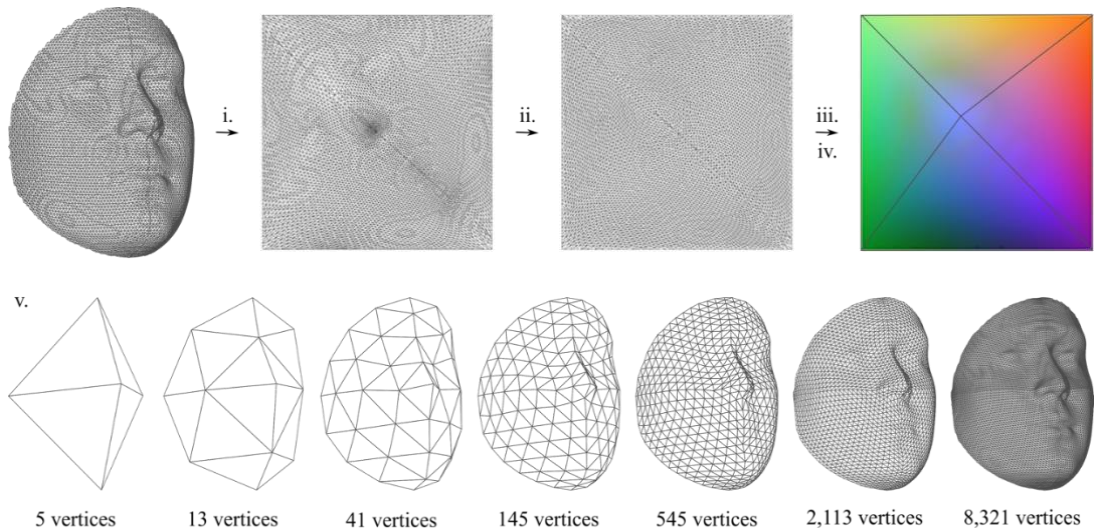

Figure 4: Steps for generating the mesh sampling scheme. The result of step (iii) is displayed as an RGB image where different colors represent different dimensions (R = x-values, G = y-values, B = z-values).

v. Finally, the meshes generated in step (iv) are reformed to represent the original facial shape by linear interpolation over the x-, y- and z-values in the output arrays of step (iii).

Since all faces in the database are represented on the same topology, the first two steps are executed only once on a canonical template mesh. Then the Euclidean representation (step (iii)) is computed for every individual in the dataset and used to reconstruct the original shapes in step (v). These steps are executed once, as a preprocessing step before neural training.

*2) Fusion-Net*

The goal of the Fusion-Net is to verify whether a given face, represented by an embedding, matches a given property or not. Furthermore, the network is also expected to match an ensemble of embeddings to a set of properties by fusing multiple traits into one final matching score. To this end, we implemented a fully connected, binary classification network, which takes a concatenation of embedding spaces and a list of properties as input, and predicts whether this combination is a genuine match or an imposter (Fig 1.b). The network is trained by presenting it with both genuine and imposter combinations and uses the cross-entropy loss function [64].

During training, each embedding is presented to the network twice every epoch, once with the correct list of properties matching the face (genuine properties) and once with an incorrect list of imposter properties. The imposter properties are randomly sampled every epoch from a set of possible imposters that is generated for each individual. The set of imposters is generated by considering all other individuals in the training set and deciding whether they are an imposter based on the trait of interest. For binary traits (sex, GB), every property that is from a different class is considered an imposter. For sex this means the opposite sex will be selected as the imposter and for genomic background any set of components that has a different sign along at least one of the first four SUGIBS dimensions can be selected.

$$I_k^{sex} = \{x: P_x \neq P_k\} \quad (3)$$

$$I_k^{GB} = x: \exists i: P_x^i \neq P_k^i \quad (4)$$

For continuous traits, the difference between the true property and the imposter should be above a given threshold to ensure they are sufficiently distinct. These thresholds are based on the distance thresholds T used in the triplet mining for the GML.

$$I_k^{cont.} = \{x: |P_x - P_k| > T\} \quad (5)$$

Finally, when combining several traits, any individual that is considered an imposter for at least one of the traits is added to the set.

$$I_k = I_k^{sex} \cup I_k^{age} \cup I_k^{BMI} \cup I_k^{GB} \quad (6)$$

*C. Training and Evaluation Strategy*

In order to evaluate our models, the dataset is divided into 10 folds (i) for a cross-validation. In each fold, 10% of the data is devoted to the test set (test$_i$), and the remaining is used for training the 2-stages of our pipeline. Hence, the training data is further split into two non-overlapping partitions train$_i^1$ (60%) and train$_i^2$ (40%), for training the first (GML) and the second stage (Fusion-Net) of the pipeline, respectively. The following experiments are designed to measure and compare the effect of our proposed GML and Fusion-Net in the context of a Face-to-DNA biometrics system:

i. The First experiment replicates the currently existing linear approach and will be considered as the baseline throughout this study. PCA is initially applied to the full facial meshes to obtain a 20-dimensional embedding space, which captures 95.3% of the data, using train$_i^1$. Then, train$_i^2$ is further divided into two partitions, to be used independently for classifier and score fuser training. Binary, linear SVM classification is used to predict sex and each GB component. Similarly, SVM linear regression is used for regressing age and BMI. Classification scores are calculated from the signed distance between each sample to the SVM decision boundary. Regression scores are signed values calculated from T- (|predicted-ground truth|), in which T refers to the distance threshold of 10 for age and 2 for BMI. Then, a Naive Bayes score fuser is trained to fuse scores from all properties into one matching score similar to [2].

ii. The second experiment is designed to measure the influence of the GML stage on the overall performance. In this setup, the traditional PCA dimensionality reduction is replaced by the concatenation of the GML outputs for each property. GML encoders for sex, age and BMI each provide us with four-dimensional embedding vectors and the GB outputs provides an additional embedding of eight dimensions. Hence, a 20-dimensional concatenated vector will replace the PCA embeddings from the baseline setup (i).

iii. The next experiment focuses on Fusion-Net. The major advantage of the Fusion-Net is that it can be applied to the embedding spaces directly, hence eliminating the need for training an embedding-to-property classifier. For this reason, the last two training stages of the baseline are replaced by Fusion-Net, and the PCA embeddings are used as the input to this fuser.

iv. Finally, the combination of GML and Fusion-Net is implemented to measure the effect of our fully neural-based pipeline. The results can be compared with the linear baseline and the other intermediate experiments.

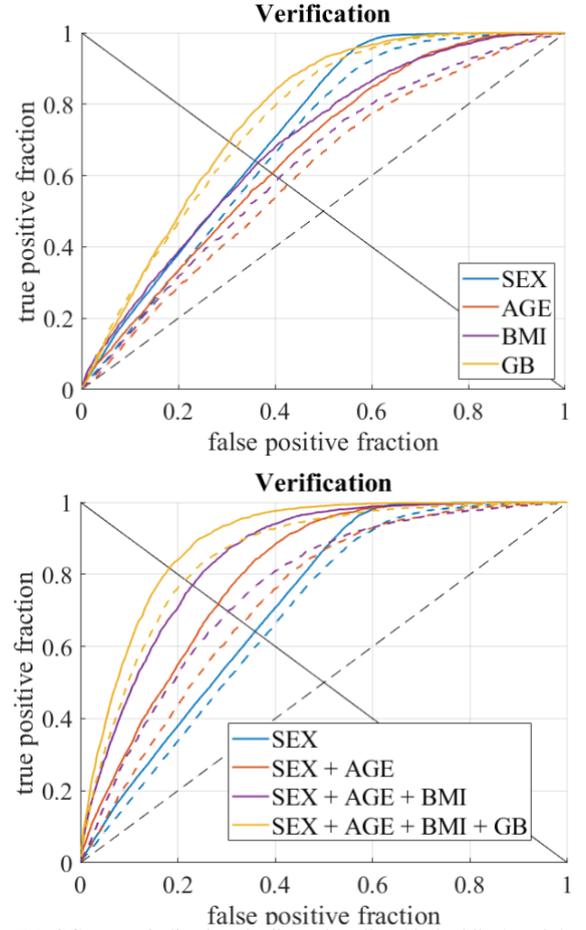

Figure 5: ROC curves indicating the linear baseline (dashed line) and the neural pipeline (full line) performances

Each of these designed architectures are evaluated in seven independent runs, with a different amount of properties included. First, the performance is investigated for each property separately. Then, age, BMI and GB are gradually added to sex in order build stronger multibiometric systems. The final system in which all properties are used to train the Fusion-Net is expected to obtain highest verification results.

## IV. RESULTS AND DISCUSSION

Matching a known person to an unidentified DNA sample can be done in different ways. The most accurate approach is DNA profiling, in which multilocus genotypes are compared to other multilocus genotypes to determine match probabilities. However, in some cases, the DNA sample of a person of interest is not available. An alternative approach in such cases is to predict the phenotype based on the DNA which is called molecular photofitting or DNA phenotyping. Due to many unknowns in the facial effects of both genetic and non-genetic factors on facial morphology, predicting faces from DNA has not been successful so far. To tackle this problem, the method we are using, which was first introduced in [2], creates an intermediate latent space to which the primary identifiers are projected. The projection of identifiers into that space are then matched against each other. In this work, the primary identifiers, which are defined as characteristics that can reliably define a person's identity, are facial shape and DNA; and the latent space is the space of properties inferred from face or DNA, which are sex, age, BMI, and GB. Since different types of identifiers (modalities) are being matched in this approach, the accuracies are expected to be inferior compared to DNA profiling. However, it is important to note that the properties we are investigating are known as soft biometrics, meaning that they do not carry sufficient information for identification purposes. These soft traits, if combined together or accompanied with primary identifiers, can improve the verification performances.

Biometric verification is a one-to-one comparison, which evaluates whether a given combination of an embedding and a property is either a match or a no-match, by testing the matching score against a threshold. The results are shown on a receiver operator characteristic (ROC) curve, which plots the true positive (TP) rate against the false positive (FP) rate for a decreasing threshold. A large area under the curve (AUC) and low equal error rate (EER) indicate better performance. Sensitivity is the probability that a positive example is correctly classified as positive, while the specificity indicates the chances of correctly denying a negative example.

The individual and cumulative verification curves, for the baseline and our neural-based biometric system are shown in Fig. 5. It shows that our pipeline outperforms the baseline, for all runs. We observe that the verification systems are sensitive, meaning they can correctly accept positive samples, but not specific and thus are unable to correctly deny all negative examples. As the ROC curves are pushed upwards and leftwards, both sensitivity and specificity increase in the neural-based pipeline compared to the linear baseline. Moreover, for both linear and non-linear pipelines, the verification specificity increases as a result of fusing more properties into the system. Therefore, it is of interest to investigate the effect of injecting more properties, e.g. texture driven attributes such as hair, skin and eye color, to the recognition system. In addition, inspecting the behavior of individual properties can suggest that the stronger properties, such as sex and GB, can increase the sensitivity of the system. Hence, investing in strong properties and adding them to the system can increase the sensitivity and specificity at the same time.

In order to confirm the contribution of GML and Fusion-Net individually, Table I provides the EER and AUC for each of the four experiments explained in section C. When comparing the performance of PCA + Fusion-Net with the baseline, the boosted performance for all runs illustrates the superiority of Fusion-Net to its linear Naïve Bayes competitor. A similar trend is observed in comparison between GML + Naive Bayes with the baseline. A closer inspection of the results for the full neural-pipeline reveals that the performance of the multi-biometric systems is either on par with or better than partially neural experiments, and the best overall performance is achieved by combining the GML with the Fusion-Net.

## V. CONCLUSION

This paper introduces a non-linear neural based alternative to the state-of-the-art face-to-DNA biometric system using spiral convolutional networks. The proposed pipeline consists of two major building blocks, GML and Fusion-Net, each contributing to the boosted performance compared to the baseline. The GML block employs spiral convolutional operators for metric learning, in contrast to the already existing generative models [36]. On top of that, we use a novel sampling method to obtain several resolutions. To learn a low dimensional semantic representation of the facial meshes, the designed spiral encoders are trained by the triplet loss function. We rebuilt the triplet selection strategy to cope with continuous properties. The second step of our proposed pipeline deploys Fusion-Net, a non-linear biometric fuser which avoids obtaining scores prior to the fusing. The performance of the final multi-biometric system trained with all properties indicates that the combination of GML and Fusion-Net improves the verification accuracy. We plan to further improve the performance by modifying the architecture of the Fusion-Net, in order to facilitate transfer learning between individual and cumulative runs. Moreover, developing a part-based GML for computing local embeddings is another foreseen extension to the current neural-based system.

TABLE I: MEAN VALUE OF AUC AND EER FOR THE VERIFICATION CURVES, AFTER 10-FOLD CROSS VALIDATION

|  |  | Sex | Age | BMI | GB | Sex + Age | Sex + Age + BMI | Sex + Age + BMI + GB |
|---|---|---|---|---|---|---|---|---|
| (i) PCA + Naïve Bayes | AUC | 0.69 ± 0.015 | 0.61 ± 0.018 | 0.63 ± 0.016 | 0.75 ± 0.015 | 0.73 ± 0.014 | 0.76 ± 0.010 | 0.85 ± 0.010 |
|  | EER | 0.38 ± 0.013 | 0.43 ± 0.013 | 0.41 ± 0.013 | 0.32 ± 0.015 | 0.33 ± 0.014 | 0.30 ± 0.011 | 0.22 ± 0.011 |
| (ii) PCA + Fusion-Net | AUC | 0.69 ± 0.022 | 0.66 ± 0.023 | 0.66 ± 0.023 | 0.77 ± 0.010 | 0.78 ± 0.015 | 0.82 ± 0.022 | 0.86 ± 0.016 |
|  | EER | 0.37 ± 0.019 | 0.40 ± 0.016 | 0.39 ± 0.019 | 0.29 ± 0.010 | 0.29 ± 0.010 | 0.26 ± 0.024 | 0.21 ± 0.019 |
| (iii) GML + Naïve Bayes | AUC | 0.71 ± 0.014 | 0.68 ± 0.019 | 0.63 ± 0.015 | 0.70 ± 0.011 | 0.80 ± 0.016 | 0.82 ± 0.013 | 0.88 ± 0.009 |
|  | EER | 0.37 ± 0.012 | 0.38 ± 0.014 | 0.40 ± 0.012 | 0.37 ± 0.007 | 0.28 ± 0.018 | 0.26 ± 0.013 | 0.19 ± 0.016 |
| (iv) GML + Fusion-Net | AUC | 0.72 ± 0.020 | 0.66 ± 0.044 | 0.69 ± 0.023 | 0.77 ± 0.010 | 0.80 ± 0.024 | 0.85 ± 0.021 | **0.89** ± 0.016 |
|  | EER | 0.36 ± 0.014 | 0.39 ± 0.031 | 0.36 ± 0.018 | 0.30 ± 0.007 | 0.28 ± 0.022 | 0.23 ± 0.021 | **0.18** ± 0.020 |

ACKNOWLEDGMENT

This work was supported by grants from the Research Fund KU Leuven (BOF-C1, C14/15/081), the Research Program of the Fund for Scientific Research - Flanders (Belgium; FWO, G078518N), the US National Institutes of Health (1-RO1-DE027023) and the US National Institute of Justice (2014-DN-BX-K031, 2018-DU-BX-0219).